\documentclass[conference]{IEEEtran}
\IEEEoverridecommandlockouts
\usepackage{cite}
\usepackage{amsmath,amssymb,amsfonts}
\usepackage{algorithmic}
\usepackage{graphicx}
\usepackage{textcomp}
\usepackage{xcolor}
\usepackage{hyperref}
\def\BibTeX{{\rm B\kern-.05em{\sc i\kern-.025em b}\kern-.08em
    T\kern-.1667em\lower.7ex\hbox{E}\kern-.125emX}}
\begin{document}

\title{Voice-Based Smart Assistant System for Vehicles using RASA}

\author{\IEEEauthorblockN{1\textsuperscript{st} Aditya Paranjape}
\IEEEauthorblockA{\textit{Information Technology} \\
\textit{Pune Institute of Computer Technology}\\
Pune, India \\
adifeb24@gmail.com}
\and
\IEEEauthorblockN{2\textsuperscript{nd} Yash Patwardhan}
\IEEEauthorblockA{\textit{Information Technology} \\
\textit{Pune Institute of Computer Technology}\\
Pune, India \\
yash23pat@gmail.com}
\and
\IEEEauthorblockN{3\textsuperscript{rd} Vedant Deshpande}
\IEEEauthorblockA{\textit{Information Technology} \\
\textit{Pune Institute of Computer Technology}\\
Pune, India \\
vedantd41@gmail.com}
\and
\IEEEauthorblockN{4\textsuperscript{th} Aniket Darp}
\IEEEauthorblockA{\textit{Information Technology} \\
\textit{Pune Institute of Computer Technology}\\
Pune, India \\
darpaniket10@gmail.com}
\and
\IEEEauthorblockN{5\textsuperscript{th} Dr. Jayashree Jagdale}
\IEEEauthorblockA{\textit{Information Technology} \\
\textit{Pune Institute of Computer Technology}\\
Pune, India \\
jbjagdale@pict.edu}
}
\maketitle

\begin{abstract}
Conversational AIs, or chatbots, mimic human speech when conversing. Smart assistants facilitate the automation of several tasks that needed human intervention earlier. Because of their accuracy, absence of dependence on human resources, and accessibility around the clock, chatbots can be employed in vehicles too. Due to people's propensity to divert their attention away from the task of driving while engaging in other activities like calling, playing music, navigation, and getting updates on the weather forecast and latest news, road safety has declined and accidents have increased as a result. It would be advantageous to automate these tasks using voice commands rather than carrying them out manually. This paper focuses on the development of a voice-based smart assistance application for vehicles based on the RASA framework. The smart assistant provides functionalities like navigation, communication via calls, getting weather forecasts and the latest news updates, and music that are completely voice-based in nature.
\end{abstract}


\section{Introduction}
A chatbot is a programmed tool that can mimic human conversations via text or voice. Chatbots can be either stateless or stateful. Stateless chatbots consider each conversation as a new one, while stateful chatbots remember past interactions and can use that data to give more personalized responses.
Voice is the fastest way for a modern digital business to communicate with its customers. Voice-enabled chatbots outperform emails, call centers, and accessing support sections on a company's website. Speech AI software has the potential to revolutionize the way businesses communicate with their customers. Voice-recognition chatbots can improve the customer experience for both the customer and the business.
RASA, an open-source framework, is best suited for creating chatbots that can provide a more human touch to interactions. It is made up of two key components, namely RASA NLU and RASA Core. RASA NLU is a tool for natural language processing that is open-source and assists chatbots in comprehending user input. It does this by categorizing user intents (what the user is asking), extracting entities from user input (such as names, dates, and locations), and assisting chatbots in understanding user input. RASA Core is a framework for chatbots that incorporates machine learning for managing dialogues. It accepts input from RASA NLU and predicts the most probable action in line using a probabilistic model (like an LSTM neural network) rather than if/else conditional lines of code. Reinforcement learning is leveraged by RASA Core to ensure enhancement in the course of prediction for the next best action. Overall, RASA is an excellent conversational AI tool that is adaptable and customizable. It is a great choice for businesses that want to improve the customer experience. This paper aims to study and create a voice-based smart assistant using the RASA framework, serving as an assistant to the user while driving a vehicle. This assistant targets drivers who wish to perform functions like playing music, setting up navigation, getting news and weather updates, and placing phone calls. The assistant will improve its efficiency by getting more real-world data from its users.

\section{Related Work}
    A chatbot is a computer program that communicates with humans and other bots by acting as an intermediary agent. Broadly, there exist three types of chatbots; contextual, menu\_button-based, and keyword recognition-based \cite{1,2,3}. AIML (Artificial Intelligence Markup Language) is a markup language used to develop distinctive chatbots. Consequently, they are very popular when it comes to choosing the type of chatbot that is to be built\cite{4}.
    Conversation systems are divided into three main categories namely agents which answer questions, dialogue agents which are task-oriented, and social bots. ReasoNet is used in the approach by Agents, which answer questions. For the Task-oriented Dialogue Agents, state-of-the-art approaches are described using two methods, end-to-end (E2E) learning and advanced reinforcement learning techniques. The exploration of a fully data-oriented stage of conversational responses in the form of encoder-decoder or seq2seq models has been started by researchers\cite{5}.
    
    Recent advancements in the field of Conversational AI architecture development have shown that Bi-directional LSTM produces the best results. Dialogue Management is responsible for the actions of the Conversational Agent and mapping inputs to appropriate outputs. Natural Language Generation (NLG) is a subdomain of Natural Language Processing (NLP), that includes the generation of outputs according to given queries in the natural language \cite{6}. Several studies were analyzed, and it was discovered that many chatbots lacked emotion as a component of input, which may be quite useful in providing even more suitable outcomes \cite{7}.
    It is seen that chatbots developed using retrieval-based models respond to queries from the training dataset using proper language and spelling. The system, however, is unable to respond to queries not contained in the training dataset, and the responses can occasionally seem forced\cite{8}.


    RASA Conversational AI assistant is built upon natural conversations. This is done with the help of contextualization. The assistant is said to have mainly two components; the RASA NLU and RASA Core \cite{9,10}. The RASA NLU is also referred to as the ear of the assistant as it handles the input while the RASA Core is referred to as the brain of the assistant because it makes decisions based on the input \cite{11}. Their goal is to provide non-specialist software developers with access to machine-learning-based dialogue management and language interpretation. Language comprehension and dialogue management in RASA are completely independent. As a result, trained dialogue models may be employed in multiple languages and RASA NLU and Core can be used independently of one another. RASA NLU consists of loosely connected modules that integrate a large number of machine learning and natural language processing components in a unified API\cite{12}.
    
    RASA has various application fields, including Intelligent Customer Service in the disease detection domain, which combines users' conversational records and existing data with semantic analysis\cite{13}. Taiwan Disease Control Center proposed and utilized a system that provided 99\% accurate disease information to users instantly\cite{14}.
    A conversational question-and-answer (CQA) system was created, and by determining the question's intent and locating the solution in a knowledge base, the system may respond to inquiries made in natural language. The limitations of the system include the fact that the system is only able to answer questions that are in the knowledge base\cite{15}.
    Intelligent virtual laboratory developed with NLP (Natural Language Processing) and machine learning models with Artificial Intelligence (AI) powered RASA framework for virtual assistant creation in this study. Chatbot is set up on the RASA chatbot server, which is a machine learning model framework to build chatbot using RASA NLU and Core. With the help of these virtual assistants, students can solve their problems faced during an experiment \cite{16}. A similar implementation involving the RASA framework has been used to create a college inquiry chatbot and a question clarification virtual assistant to aid students in resolving their doubts in another application \cite{17, 18}. RASA has also been employed to develop a chatbot designed to assist in the generation of academic schedules \cite{19}. RASA has found its application in the locomotive industry too. A virtual assistant created using RASA is used for automating information exchange in a car \cite{20}.
    
    This paper proposes a voice-based smart assistant system for vehicles leveraging the benefits of the RASA framework.

\section{Methodology}
\subsection{RASA}
RASA is an open-source framework used to build text-based and voice-based chatbots. RASA chatbots are capable of contextualization and can devise their response based on the surrounding context. As a result, they are said to work at Level 3 of conversational AI. Transformer Embedding Dialogue Policy(TED), which is the default machine learning-based dialogue policy used by RASA, is the algorithm which the models are trained over. It is also the default machine learning-based dialogue policy used in the framework. TED is especially useful for handling unforeseen situations, and can efficiently handle user input even if the conversation goes off track. To choose which dialogues to selectively pay attention to and ignore while making predictions, TED uses a transformer architecture. The TED policy takes three main inputs, viz. the user's utterance, predicted previous action, and slots. The transformer accepts this information after it is featured and concatenated.

In this paper, a RASA framework is used to create a smart assistant to assist drivers while on the road. Voice-based functionalities using speech-to-text technologies are added to ensure that drivers need not look at the screen while driving. Whenever the user speaks, the speech-to-text functionality will convert whatever the user has said into text and send it to the NLU. In the NLU, entity, and intent recognition will take place, where the keywords of the sentences will be picked and contextualized. Based on these words, the core will decide the relevant response and call the respective API or reply to the user based on its knowledge. While replying to the user, the text-to-speech functionality will convert the text back to audio to enhance the user experience and reduce the need to look away from the road while driving. It makes use of intents, entities, stories, and rules to build up and understand various possible scenarios.

\begin{enumerate}

\item Intents: Intent defines the purpose or aim of speaking or performing a certain action. (e,g. expressing disappointment, bidding farewell).

\item Entities: Entities are the keywords that help the model make relevant recommendations, which are obtained through the user's input. (e.g. location, name of song, color).

\item Stories: Stories contain a format of training data for the model which comprises the interaction of the user with the bot. The bot's responses are defined by a set of ordered actions, whereas the user's messages are interpreted as intents and entities.

\item Rules: Rules provide a flow of actions where a specific next action is always predicted by a triggering condition. (e.g. handling errors, similar stories)
\\
\end{enumerate}
RASA has two main components: RASA NLU and RASA Core.
\begin{enumerate}
    \item RASA NLU: RASA NLU refers to the Natural Language Understanding component which examines the user's utterance, classifies the intent, and abstracts the entities.
    \item RASA Core: RASA Core contains the engine that takes decisions regarding the next actions to be undertaken according to the context of the conversation. It accepts input from the user and comes up with a response with the help of pipelines in RASA. \\
\end{enumerate}

\begin{figure}[ht]
    \centering
    \includegraphics[width=0.5\textwidth]{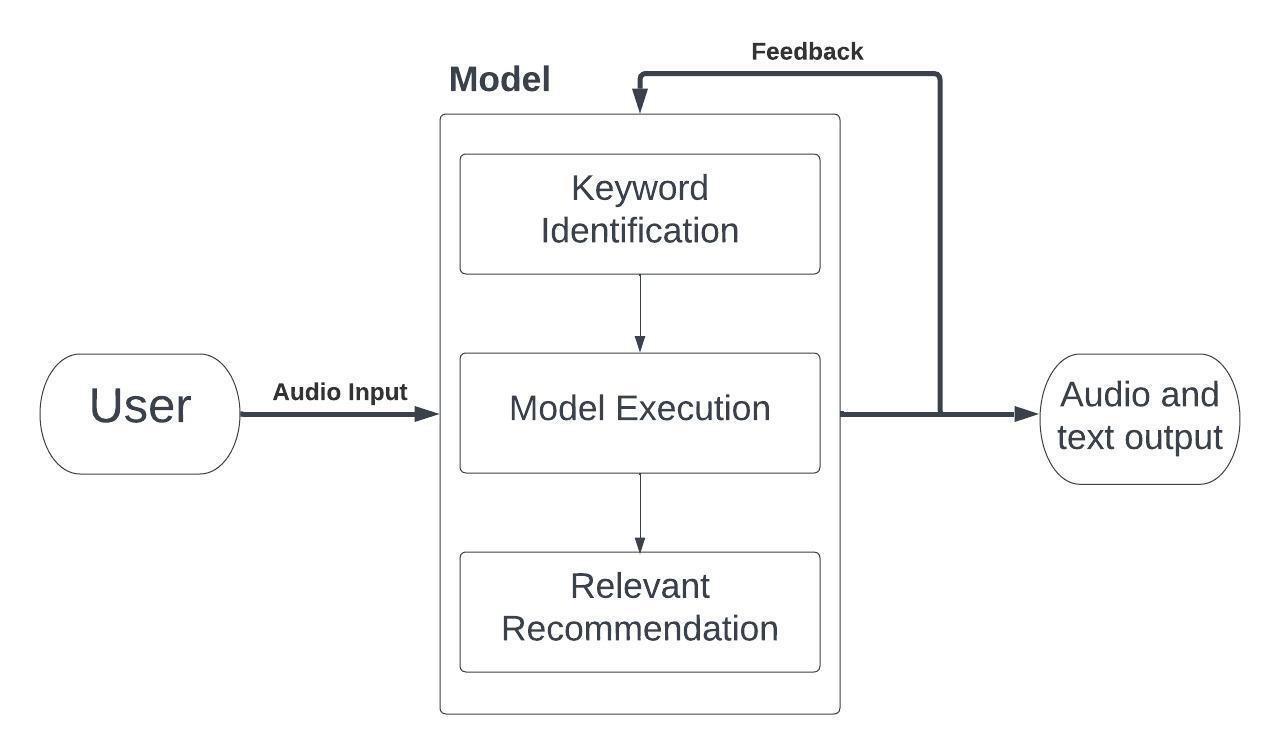}
    \caption{System Architecture for the Voice-Based Smart Assistant}
    \label{fig:sysarch}
\end{figure}

\autoref{fig:sysarch} shows the system architecture of the smart assistant. The system processes the audio input obtained from the user by extracting the keywords to identify the intents and triggering necessary actions. The pre-trained model runs to give a relevant response to the user query.

\subsection{Modules}
\begin{enumerate}

    \item Voice Input \& Output \\
    The input and output process of the bot is entirely voice-based, achieved using the open-source text-to-speech (TTS) library 'pyttsx3' and the speech-to-text (STT) library 'speech-recognition'. The pyttsx3 engine is initialized, and properties such as voice, speaking rate, and volume are set. 
    Once the trigger word, in this case, 'coffee', is detected by the program, the voice bot starts execution and gets input from the user. 
    The user's input is then sent to the RASA webhook as a POST request to obtain the desired output.
    \\
    \item RASA Webhook \\
    In RASA, a webhook is a way to connect the chatbot to an external web service, allowing it to fetch or send data to that service. Webhooks are used to perform actions outside of the chatbot, such as looking up information in a database or sending a notification to a user. 
    In this system, POST requests are sent to the webhook endpoint along with the message in the form of JSON. The response to this request will contain the output of the bot. 
    \\
    \item General Interaction \\
    This module is used to greet the user when the user initiates and terminates the assistant. The greet intent is identified using words like hello, hi, hey, etc. The utter\_greet action is performed by the assistant to reply and greet the user. If the user says goodbye to the bot, the bot reciprocates the same and the bot is stopped temporarily.  \\
    \item News \\
    The main objective of the news module is to provide the latest news headlines to the user, which are fetched using the open-source NewsAPI. A story is designed to deliver the headlines to the user, which is initiated once the intent is identified correctly. The news intent is identified using words like news, headlines, and related key terms. The bot then conveys to the user that news headlines are being fetched. The action calls the NewsAPI with the appropriate parameters, and results are obtained in JSON format. The bot asks the user whether or not they would like to listen to more news headlines. If the user affirms, the API call is repeated, and other news headlines are provided to the user.   \\


    \item Weather \\
    The objective of the weather module is to provide the current weather data and the weather forecast for the next day for a given user-specified location. WeatherAPI has been used to fetch the data. The weather data includes temperature, humidity, pressure, wind speed and direction, cloud cover, and more. A story is designed to deliver the weather data to the user, which is initiated once the intent is identified correctly. The intent is identified using words like the weather, weather updates, today's weather report, etc. The action prompts the user to specify the location and the next actions are used to verify that the location interpreted by the assistant is correct. The intent affirm is identified when the user responds positively to verify that the location understood by the assistant is correct and the Weather API is called using appropriate parameters. The relevant information from these results is conveyed to the user using the voice module.     \\

    \item Navigation \\
    The objective of the navigation module is to provide the route to the specified destination using the Google Maps API. A story is designed to display the route to the destination to the user, which is initiated once the intent is identified correctly. The navigation intent is identified using words like maps, directions, open maps, etc. The action prompts the user to specify the destination. The intent for input location is identified if the user speaks the name of a real location. Subsequently, verification actions are used to confirm that the location is interpreted correctly by the assistant. On positive affirmation from the user for the confirmation action, the API call for Google Maps displays the route to the destination from the current location.     \\

    \item Music \\
    The objective of the music module is to play the song requested by the user, which is fetched using the Spotify API. A story is designed to play a specific song for the user, which is initiated once the intent is identified correctly. The intent is identified using words or phrases like music, song, let's have some music, etc. The bot then asks the user to specify the song and records the user's input. The next actions verify from the user that the assistant has heard the correct name of the song and on positive affirmation, the Spotify API is called with the appropriate parameters, and the requested song is played.  \\

    \item Communication \\
    The objective of the communication module is to place a phone call to the person requested by the user. A story is designed to place a phone call to the user-specified contact, which is initiated once the intent is identified correctly. The intent is identified using the words call, phone, contact, etc. The assistant then prompts the user to provide the input for the contact name. The next actions are used by the assistant to verify from the user that the assistant has identified the name correctly. The phone call is then placed to the appropriate contact. \\

    \begin{figure}[ht]
    \centering
    \includegraphics[width=0.5\textwidth]{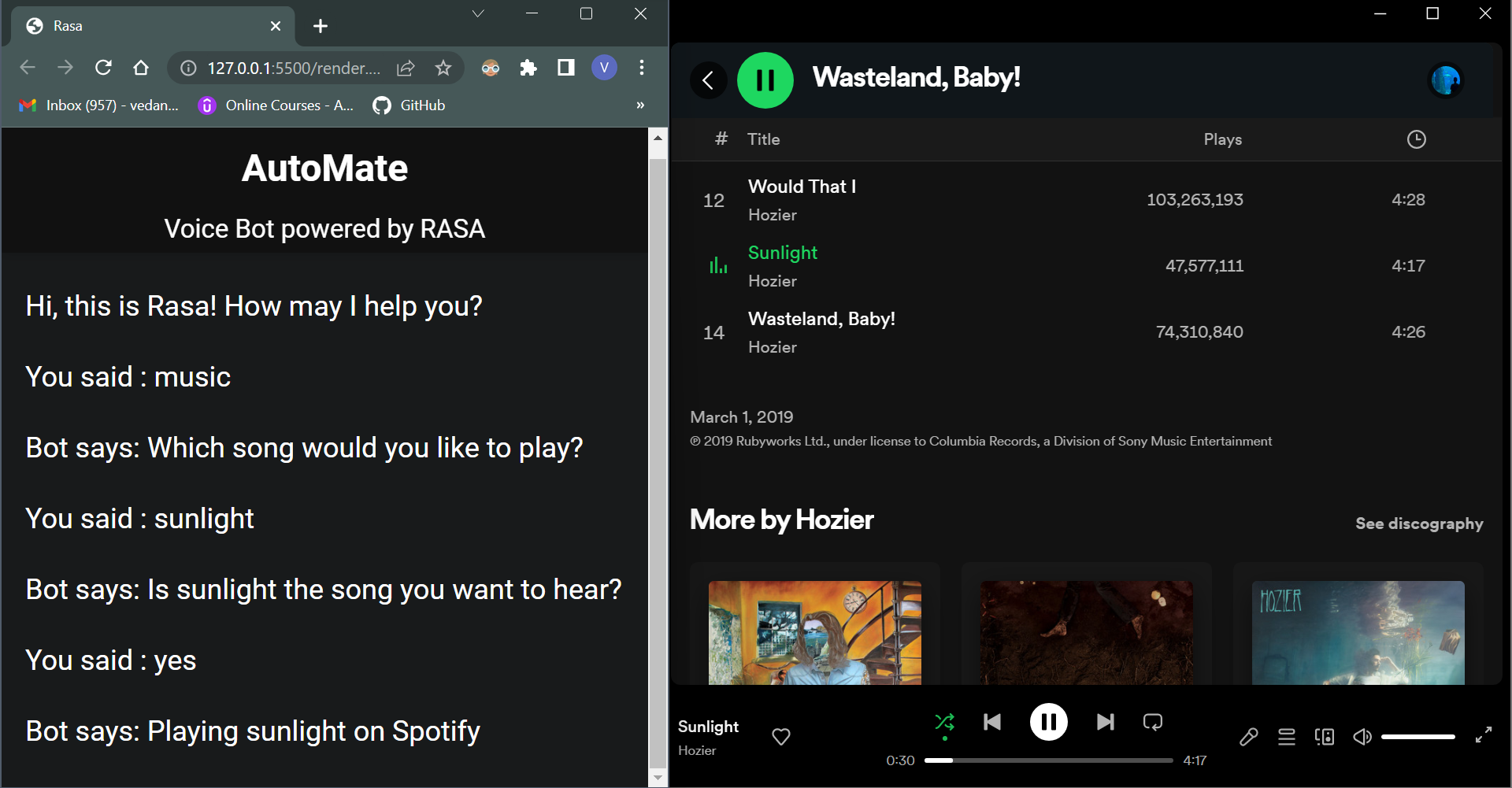}
    \caption{Music Module Output}
    \label{fig:music}
    \end{figure}

     \autoref{fig:music} depicts the output of the music module in a visual form. Although the system is predominantly voice-based, a basic frontend functionality has been provided to the user.
    
\end{enumerate}

\section{Experimentation and Results}

\subsection{Experimentation Setup}
Hyper-parameters are used to fine-tune the model. The hyper-parameters are defined in the pipeline. In RASA, a pipeline refers to a series of components that are used to process and transform user inputs, such as text messages or voice commands, into structured data that can be used by the framework to generate appropriate responses or take actions. WhiteSpaceTokenizer tokenizes the input text based on whitespace, splitting it into individual words or tokens. DIET classifier is a neural network-based classifier that uses a combination of pre-trained word embeddings and sequence modeling to classify text. It is trained using the supervised learning approach and optimized with the DIET algorithm. The "epochs" parameter specifies the number of training epochs for the model, and "constrain\_similarities" enforces similarity constraints during training. The order of the components in the pipeline is important since each component takes the output of the previous component and uses it as input. The following pipeline is used to create this voice-based chatbot.

\begin{figure}[ht]
    \centering
    \includegraphics[width=0.5\textwidth]{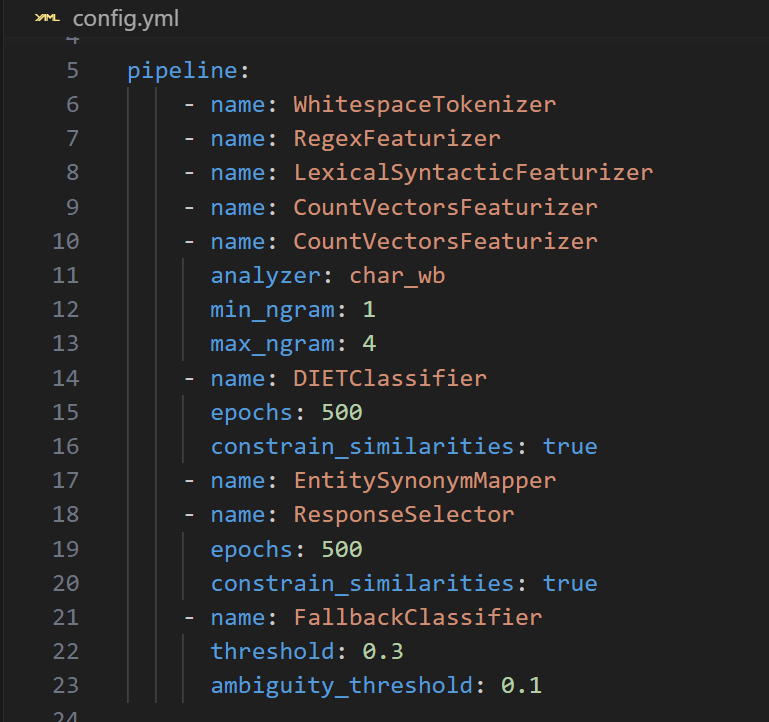}
    \caption{Pipeline for the Voice-Based Smart Assistant}
    \label{fig:myimage}
\end{figure}

In \autoref{fig:myimage}, it can be seen that the pipeline performs a series of text processing steps,
featurization, and classification to understand and respond to user input. It incorporates
various techniques, such as tokenization, feature extraction, and neural network-based
classifiers, to improve the accuracy and performance of the text classification task.

\subsection{Results}
The results of this study can be divided into 3 main components: average response time for each action, average response time for each module, and accuracy of intent identification.
The time required to transmit the query, process it by the computer, and then communicate the response is included in the response time. The response time of an
interactive system is typically used to assess its performance. The average response times for a specific action were determined based on a sample of 100 instances. Actions are divided into 3 main types, viz. Intent Identification, Input \& Confirmation, and API Call \& Output.
\autoref{tab:table1} shows the average response time (in seconds), for each type of action. For example, the time required for the bot to respond to the user to perform intent identification is 0.08 seconds.

\renewcommand{\arraystretch}{1.5}

\begin{table}[ht]
\centering
\caption{Average response times of the bot for each type of action}
\label{tab:table1}
\begin{tabular}{|c|c|}
\hline
Type of Action        & Avg Response Time (in seconds) \\ \hline
Intent Identification & 0.08                               \\ 
Input \& Confirmation & 2.4                                \\ 
API Call \& Output    & 4.6                                \\ 
\hline
\end{tabular}
\end{table}

The average response times for a specific module were determined based on a sample
of 100 instances. This includes an average of the response times of the 3 main actions present in a module. Modules are divided into 6 main types, viz. General Interaction, Music, Navigation, Communication, Weather \& News. \
\autoref{tab:table2} shows the average response times of the bot (in seconds) for every module. For example, while accessing the news module, the user can expect a response from the bot in 2.12 seconds on average.  

\renewcommand{\arraystretch}{1.5}

\begin{table}[ht]
\centering
\caption{Average response times of the bot for each module}
\label{tab:table2}
\begin{tabular}{|c|c|}
\hline
Module        &         Avg Response Time (in seconds)      \\ \hline
General Interaction        & 0.06                           \\
Music                      & 2.62                           \\
Navigation                 & 2.46                           \\
Communication              & 2.19                           \\
Weather                    & 2.58                           \\
News                       & 2.12                           \\ \hline
\end{tabular}
\end{table}

The automatic categorization of text data based on predefined rules is known
as intent classification. Intent classification automatically associates words or sentences with a specific intent using machine learning and natural language processing. Accuracy is the ratio of rightly classified data that a trained ML model produces, or the proportion of correct predictions to all other forecasts combined. It is a statistic for assessing the effectiveness of models in categorization tasks. 
The intents were classified correctly for 281 out of 300 instances by the proposed model, yielding an accuracy of 93.67\%. Samples of the inputs in consideration have been displayed in \autoref{tab:my-table}. 

\renewcommand{\arraystretch}{1.5}

\begin{table}[ht]
\centering
\caption{Accuracy of identifying intents}
\label{tab:my-table}
\begin{tabular}{|c|c|c|c|}
\hline
Intent      & Identified Intent & Correct Intent & Result (Y/N)  \\ \hline
Sunlight    & Song              & Song           & Y                           \\ 
Mumbai      & Location          & Location       & Y                           \\
Delhi       & Location          & Location       & Y                           \\
John        & Person            & Person         & Y                           \\
Suresh      & Location          & Person         & N                           \\
99 Problems & Song              & Song           & Y                           \\
Sachin      & Person            & Person         & Y                           \\
New York    & Location          & Location       & Y                           \\
Stan        & Song              & Song           & Y                           \\
Paris       & Location          & Song           & N 
\\ \hline
\end{tabular}
\end{table}

In the examples presented in \autoref{tab:my-table}, 8 of the 10 intents were properly identified. Suresh, as a proper noun and a regional name, is misinterpreted. Although Paris is identified as a location, the user anticipated the bot to identify Paris as a song rather than a location.

\begin{figure}[ht]
    \centering
    \includegraphics[width=0.5\textwidth]{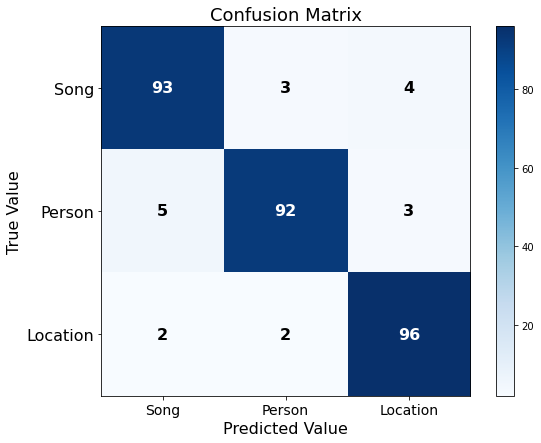}
    \caption{Confusion Matrix}
    \label{fig:fullwidth}
\end{figure}

As mentioned previously, 281 out of 300 intents were correctly identified. The confusion matrix helps in understanding the overall performance of the models and \autoref{fig:fullwidth} shows a graphical representation of the obtained results. 

\section{Conclusion}
This paper proposes a voice-based smart assistant for vehicles based on RASA. The system's goal is to reduce distractions while driving vehicles by facilitating routine driving activities via an audio channel, encouraging safe driving. The system includes frequently utilized functionalities such as general interaction, news, weather, navigation, music, and communication. The results of the research were quantified using three metrics: average response time for each action, average response time for each module, and accuracy of intent identification. From the user's perspective, the most important metric is the accuracy of intent identification, which comes out to be 93.67\%. However, the model has some limitations, including deteriorated performance in a noisy environment, difficulty understanding regional accents, and limited training data. The future scope of this study may include the implementation of multi-lingual models to aid in the understanding of multiple languages, integration with hardware via the Internet of Things, and the implementation of additional functionalities such as voice-automating vehicle controls such as changing the radio volume, locking the car doors, enabling wipers, and controlling the air conditioning, to name a few.







\bibliographystyle{plain}

\end{document}